\documentclass[journal]{IEEEtran}

\usepackage{graphicx}%
\usepackage{multirow}%
\usepackage{amsmath,amssymb,amsfonts,bm}%
\usepackage{amsthm}%
\usepackage{mathrsfs}%
\usepackage{xcolor}%
\usepackage{textcomp}%
\usepackage{manyfoot}%
\usepackage{booktabs}%
\usepackage{algorithm}%
\usepackage{algorithmicx}%
\usepackage{algpseudocode}%
\usepackage{listings}%
\usepackage{gensymb}
\usepackage{diagbox}
\usepackage{placeins}
\begin{document}
%
\title{RFM-Pose: Reinforcement-Guided Flow Matching for Fast Category-Level 6D Pose Estimation}
%
%
%

\author{Diya~He,
        Qingchen~Liu,~\IEEEmembership{Member,~IEEE,}
        Cong~Zhang,        Jiahu~Qin,~\IEEEmembership{Senior~Member,~IEEE}

\thanks{*Corresponding author: Jiahu Qin (e-mail: jhqin@ustc.edu.cn).}%
\thanks{D. He, Q. Liu, C. Zhang and J. Qin are with the Department of Automation, University of Science and Technology of China, Hefei 230027, China (e-mail: hodia@mail.ustc.edu.cn; qingchen\_liu@ustc.edu.cn; Cong\_Zhang@ustc.edu.cn; jhqin@ustc.edu.cn).}
}

\maketitle

\begin{abstract}
Object pose estimation is a fundamental problem in computer vision and plays a critical role in virtual reality and embodied intelligence, where agents must understand and interact with objects in 3D space. Recently, score based generative models have to some extent solved the rotational symmetry ambiguity problem in category level pose estimation, but their efficiency remains limited by the high sampling cost of score-based diffusion. In this work, we propose a new framework, RFM-Pose, that accelerates category-level 6D object pose generation while actively evaluating sampled hypotheses. To improve sampling efficiency, we adopt a flow-matching generative model and generate pose candidates along an optimal transport path from a simple prior to the pose distribution. To further refine these candidates, we cast the flow-matching sampling process as a Markov decision process and apply proximal policy optimization to fine-tune the sampling policy. In particular, we interpret the flow field as a learnable policy and map an estimator to a value network, enabling joint optimization of pose generation and hypothesis scoring within a reinforcement learning framework. Experiments on the REAL275 benchmark demonstrate that RFM-Pose achieves favorable performance while significantly reducing computational cost. Moreover, similar to prior work, our approach can be readily adapted to object pose tracking and attains competitive results in this setting. 
\end{abstract}

\begin{IEEEkeywords}
Category-level 6D object pose estimation, Flow matching, Proximal policy optimization.
\end{IEEEkeywords}

%
\IEEEpeerreviewmaketitle

\section{Introduction}\label{sec:intro}

Category-level 6D object pose estimation aims to recover the rotation and translation of an object instance that has never been seen during training but belongs to a known semantic category. This setting is central to vision-based robotics, virtual reality, and embodied intelligence, where object instances change frequently and precise CAD models or multi-view templates are often unavailable~\cite{Liu2024DeepLO, Wang2019NormalizedOC}. Beyond single-frame estimation, extending such methods to continuous pose tracking~\cite{9792223} is crucial for dynamic manipulation tasks. Compared with instance-level pose estimation, category-level methods must tolerate substantial intra-class variations in shape and appearance; compared with unseen-object settings, they benefit from category-level regularities but still face severe challenges~\cite{Liu2024DeepLO, Wang2019NormalizedOC}. A major difficulty of category-level pose estimation is its inherently rotational symmetry ambiguity. 

Partial observations and clutter often yield multiple plausible rotations and translations, rotational ambiguity can be further exacerbated for objects with relatively symmetric geometries. 
Early deep learning methods addressed the ambiguity caused by object symmetries through specialized loss functions or implicit representations. For instance, \cite{xiang2018posecnn} proposed the shape-match loss function to measure alignment error in 3D space, avoiding convergence issues associated with direct parameter regression. To bypass explicit symmetry labeling, \cite{sundermeyer2018implicit} learned implicit orientation embeddings that map similar views to nearby latent points, while \cite{hodan2020epos} utilized surface fragments to establish dense correspondences. However, these deterministic approaches often struggle to provide robust predictions under severe occlusion or clutter.
Recent generative approaches, particularly diffusion models, effectively handle ambiguity by modeling uncertainty through diverse hypotheses~\cite{zhang2024generative, Zhang2024Omni6DPoseAB}. However, they incur high inference costs due to the need for extensive iterative denoising and subsequent ranking of numerous candidates. Crucially, the disjoint nature of generation and evaluation prevents the model from being directly optimized using pose accuracy feedback.

\begin{figure}[t]
\centering
\includegraphics[width=0.8\columnwidth]{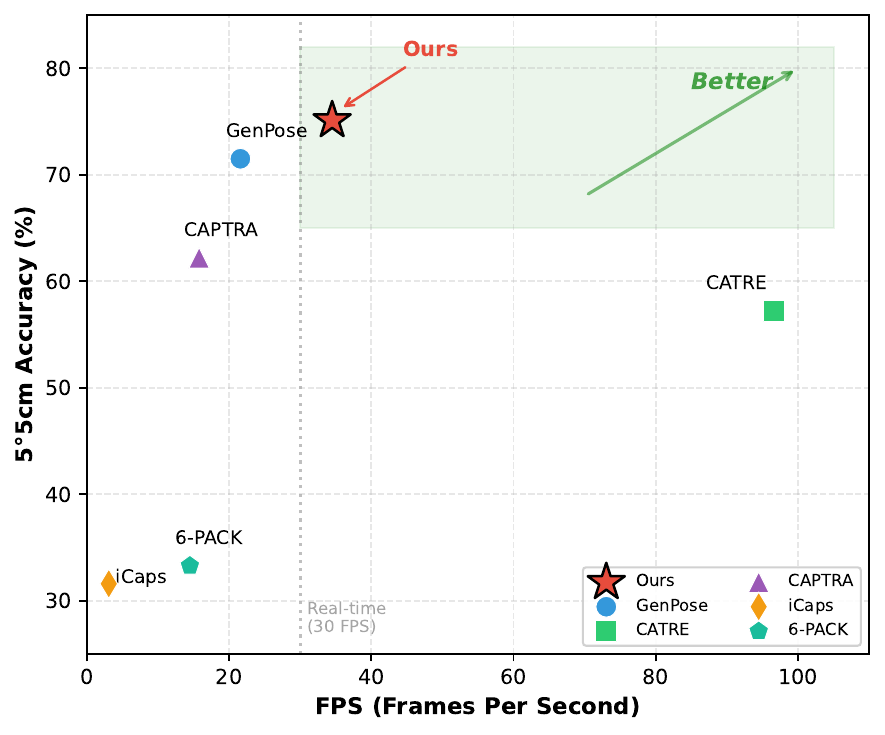}
\caption{
Speed-accuracy trade-off comparison on REAL275. Our method achieves favorable balance between inference speed and pose accuracy.}
\label{fig:speed_accuracy_comparison}
\end{figure}

To overcome these efficiency limitations, we propose RFM-Pose, a reinforced framework leveraging Flow Matching (FM) for fast and accurate category-level 6D pose estimation. 
Our approach builds upon FM~\cite{Lipman2022FlowMF, Liu2022FlowSA}, which learns a deterministic transport field to map a simple prior to the target data distribution. 
Unlike score-based diffusion, FM facilitates fast ODE-based generation with substantially fewer integration steps, making pose prediction practical for real-time applications. 
However, directly replacing score matching with flow matching introduces two distinct challenges. 
First, unlike diffusion models that naturally provide a relative likelihood signal for ranking~\cite{zhang2024generative}, FM lacks an intrinsic, reliable scoring rule for candidate selection.
Second, we observe that the pretrained FM model often yields an overly dispersed rotational distribution on $\mathrm{SO}(3)$. This arises because certain objects exhibit geometric symmetry, and FM is trained to match the velocity field rather than directly minimizing the final pose error. 
This dispersion not only complicates the subsequent ranking but also degrades the precision of the final estimation.

Due to the aforementioned restrictions, we addresses these gaps by tightly coupling fast FM generation with reinforcement learning (RL) based refinement and valuing. 
Inspired by prior work that reinterprets iterative generative sampling as sequential decision making~\cite{Janner2022PlanningWD, Chi2023DiffusionPV}, we model the FM sampling trajectory as a Markov decision process (MDP). The pretrained flow model serves as an initial policy, and each flow model sampling step is treated as policy action. 
Instead of relying on likelihood estimation, we introduce a value-based refinement mechanism where a multi-critic value network evaluates the quality of the current hypothesis for rotation and translation, and provides dense learning signals to optimize the policy via proximal policy optimization (PPO)~\cite{Schulman2017ProximalPO}. 
Crucially, by maintaining a fixed integration schedule, we preserve the high inference speed of the original FM. 
The RL update is strictly confined to refining the directionality of the flow field based on pose value feedback.
Consequently, the resulting $\mathrm{SO}(3)$ distribution becomes significantly tighter, and the value network naturally simplifies the downstream candidate selection.

At inference time, we generate a set of pose candidates using the refined policy and ranks them using the learned value network, eliminating the need for costly likelihood estimation. The retained candidates are further aggregated to produce an accurate final pose estimate. 
As illustrated in Fig.~\ref{fig:speed_accuracy_comparison}, RFM-Pose achieves a favorable speed-accuracy trade-off, attaining higher accuracy than GenPose while operating above the real-time threshold of 30 FPS.
Overall, RFM-Pose unifies fast generation and principled evaluation within a single learnable framework, flow matching provides the speed, while RL supplies the missing scoring mechanism and reshapes the pose distribution toward higher-quality modes.

Our main contributions are summarized as follows:
\begin{itemize}
    \item We present \textbf{RFM-Pose}, a fast and robust flow-matching framework for category-level 6D pose estimation, reducing the required sampling steps by an order of magnitude compared with diffusion-based generative methods. 
    \item We reformulate flow matching sampling as a Markov decision process and introduce a PPO based refinement with a multi-critic value estimator for rotation and translation, replacing the likelihood-based scoring used in diffusion models. This unified refinement and evaluation scheme both tightens the learned $\mathrm{SO}(3)$ distribution and enables faster, more accurate candidate selection.
    \item Extensive experiments on REAL275 and Omni6DPose demonstrate that RFM-Pose achieves favorable accuracy-efficiency trade-offs compare with existing diffusion model pose estimation methods, and naturally extends to pose tracking with competitive performance.
\end{itemize}

\begin{figure*}[t]
\centering
\includegraphics[width=0.85\textwidth]{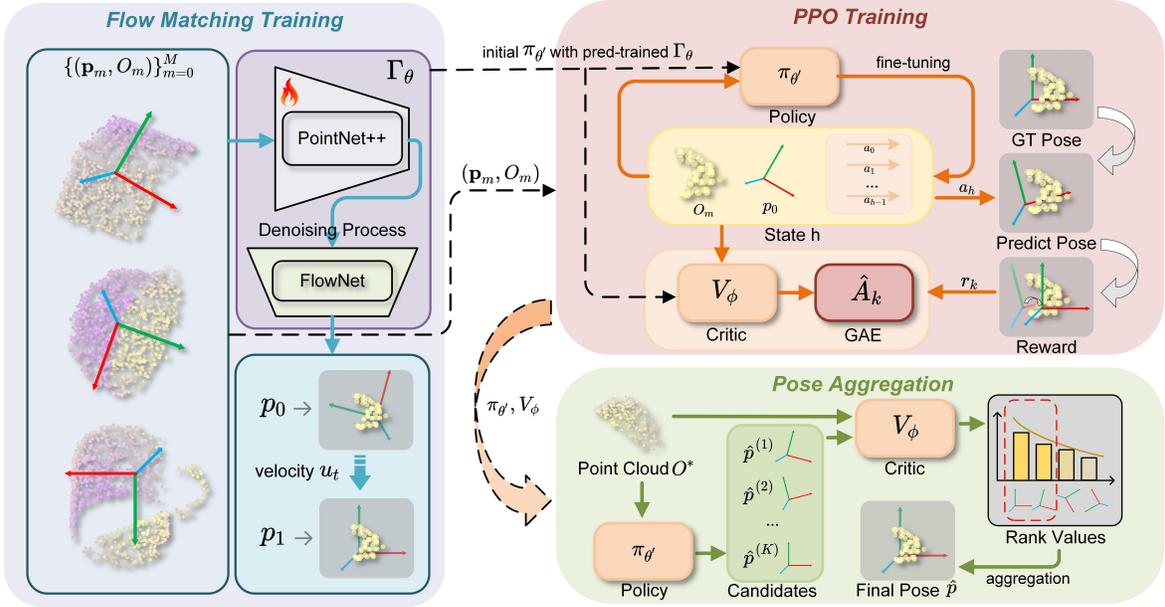}
\caption{
Overview of the RFM-Pose framework consisting of three stages. \textbf{Left:} Flow-matching model training using conditional flow-matching objective with PointNet++ feature extraction. \textbf{Right Top:} PPO-based refinement where the pretrained flow model serves as initial policy, generating pose candidates and optimizing policy and value networks with rotation/translation error rewards. \textbf{Right Bottom:} Inference stage where $K$ candidates are generated, ranked by value network(critic), and aggregated via QUEST (rotation) and weighted averaging (translation) to obtain final pose estimate.
}
\label{fig:method_structure}
\end{figure*}

\section{Related Work}\label{Related_Work}

\subsection{Category-Level Object Pose Estimation}

Category-level 6D object pose estimation has attracted increasing interest due to its ability to generalize to unseen instances within the same semantic category without relying on object-specific CAD models~\cite{Sahin2018Categorylevel6O, 10720917}. Since the introduction of NOCS~\cite{Wang2019NormalizedOC}, the problem has become a mainstream research direction. The goal is to predict the pose of arbitrary instances in a category using only partial RGB-D observations, without requiring instance-level geometry or multi-view rendering~\cite{rnek2023FoundPoseUO, Wen2023FoundationPoseU6, 10870128}. This task remains challenging due to substantial intra-class variation as well as occlusion, texture ambiguity, illumination changes, and the non-differentiability of shape alignment~\cite{Liu2024DeepLO, Wang2019NormalizedOC, zhang2024generative}.


Early approaches such as NAS~\cite{Chen2020CategoryLO} and iNeRF~\cite{yen2020inerf} optimize poses via differentiable rendering. Existing approaches include shape-prior methods~\cite{2021SGPA, Rezazadeh2023HierarchicalGN, dpod} that utilize deformable templates, and shape-agnostic methods~\cite{Di2022GPVPoseCO, Lin2022CategoryLevel6O, Zheng2023HSPoseHS, Wang2023Query6DoFLS} that directly learn geometric representations. Recent transformer-based works~\cite{lin2023vi, 2023SecondPose, 2024CatFormer, Liu2022CategoryLevel6O} further enhance rotation stability via SE(3)-equivariant representations. 
More recently, with the advent of generative models, pose estimation has increasingly been treated as distribution modeling on the SE(3) manifold~\cite{zhang2024generative, 11177583, 20236D}, among which GenPose~\cite{zhang2024generative} and 6D-Diff~\cite{20236D} explicitly model multi-hypothesis distributions through score matching, sampling multiple candidate poses and ranking them with energy-based models. These methods achieve favorable performance on benchmarks such as REAL275, yet their practical applicability is limited by the large computational cost associated with diffusion sampling. Although Omni6DPose~\cite{Zhang2024Omni6DPoseAB} introduces a large-scale dataset to advance category-level research, it does not alleviate the sampling inefficiency inherent to score-based generative models.

\subsection{Generative Models and Reinforcement Learning}

The integration of generative models and reinforcement learning has recently emerged as a powerful paradigm for embodied intelligence. As reviewed by Moroncelli et al.~\cite{Moroncelli2025Duality}, generative models offer strong priors for world modeling and multimodal representation learning, while RL enables action-driven optimization through interaction. Their combination has accelerated progress in robot learning and decision-making systems. In vision and robotics, generative models are often used as information priors for RL, providing rich structural cues and guiding exploration~\cite{Wang2023RLFM, Ahn2022SayCan, Ma2024EUREKA}. Conversely, RL can fine-tune generative models, improving their executability, adaptability, and consistency with downstream tasks.

Early works explored treating the reverse pass of a diffusion model as a parameterized policy. Ajay et al.~\cite{Ajay2022IsCG} optimized diffusion trajectories using policy gradients, while Janner et al.~\cite{Janner2022PlanningWD} introduced value-guided diffusion planning to improve trajectory quality. Diffusion Policy~\cite{Chi2023DiffusionPV} modeled robot control as a conditional diffusion process and fine-tuned it using PPO~\cite{Schulman2017ProximalPO} for high-precision manipulation. Building on this line of work, Psenka et al.~\cite{Psenka2023LearningAD} proposed Q-Score Matching, aligning the score function with the Q-function to learn diffusion-based policies directly from rewards. Moroncelli et al.~\cite{Moroncelli2025Duality} categorize these approaches under the ``RL for Generative Policies'' paradigm, emphasizing RL not only for training generative policies but also for their fine-tuning and distillation. ReinFlow~\cite{zhang2025reinflowfinetuningflowmatching} extends this idea by incorporating RL with flow matching for online policy improvement, achieving high sample efficiency and robustness.

Overall, these approaches demonstrate the effectiveness of formulating generative sampling as sequential decision-making, providing valuable insights for integrating RL-based refinement with generative models in pose estimation.

\section{Methodology}\label{Methodology}

\textbf{Task Definition.}
We aim to estimate the 6D object pose from a partially observed point cloud. 
The agent is provided with a training dataset 
$\mathcal{D}=\{ ( \mathbf{p}_m, O_m ) \}_{m=0}^{M}$, 
where $\mathbf{p}_m$ denotes the ground-truth 6D pose 
and $O_m \in \mathbb{R}^{3 \times N}$ is a partial point cloud containing $N$ points. 
The goal is to capture the underlying conditional distribution of poses given geometry. Consequently, given an unseen observation $O^{*}$ at test time, the objective is to recover its corresponding pose $\hat{p}$ by querying the learned model.

\textbf{Overview.}
To tackle the multimodal uncertainty inherent in partial observations, we propose RFM-Pose, a framework that reimagines pose estimation as a sequential decision-making process. As illustrated in Fig.~\ref{fig:method_structure}, our approach orchestrates a transition from generative modeling to reinforcement learning across three cohesive stages.

First, to establish a high-fidelity generative backbone, we train a flow-matching model $\bm{\Gamma}_{\theta}$ on $\mathcal{D}$. While powerful, purely generative models often lack direct feedback on geometric alignment quality. To bridge this gap, we reinterpret the flow sampling dynamics as a Markov decision process (MDP) defined by the state $s_h = \left\{O, \mathbf{u}_{h}, h/H\right\}$. Here, $u_0 \sim \mathcal{N}(0,I)$ represents the initial latent pose, $h$ denotes the step index within the max step $H$ and $h<H$. $\mathbf{u}_{h}=\{u_0, \dots, u_{h-1}, u_h\}$ tracks the history of predicted flow fields.

In this MDP formulation, the pre-trained flow-matching model acts as the initial policy $\pi_{\theta'}$, and each predicted flow vector $u$ is viewed as an action guiding the pose trajectory. Our core innovation lies in fine-tuning the policy to maximize the expected cumulative return, which explicitly rewards geometric accuracy:
\begin{equation}
\max_{\theta'} \mathbb{E}_{\xi \sim \pi_{\theta'}} \left[ \sum_{h=0}^{H-1} r_h \right],
\end{equation}
where $\xi$ denotes a sampling trajectory and $r_h$ combines rotation and translation rewards as detailed below. Since the integration horizon $H$ is fixed and later steps more directly reflect final pose quality, we do not discount future rewards. A multi-critic value network $V_{\phi}$ is jointly trained to estimate expected returns for variance reduction. This optimization is performed using proximal policy optimization (PPO)~\cite{Schulman2017ProximalPO}.

During inference, we leverage the optimized policy to generate a diverse set of $K$ pose results $\{\hat{p}^{(1)}, \dots, \hat{p}^{(K)}\}$. Rather than simple averaging, these candidates are critically assessed and ranked by the learned value network $V_{\phi}$, allowing us to aggregate the higher values into a final, highly accurate estimate $\hat{p}$.

\subsection{Flow Matching Training for Pose Generation}

The first challenge in estimating 6D pose from partial data is capturing the complex, multimodal distribution of valid poses. We adopt flow matching~\cite{Lipman2022FlowMF, Liu2022FlowSA} as our generative backbone, favoring its efficient sampling properties over the stochastic differential equations used in standard diffusion models~\cite{Song2020DenoisingDI}.

The intuition behind flow matching is to learn a deterministic optimal transport path that smoothly transforms a simple Gaussian prior into the complex target distribution $P_{\text{data}}(p \mid O)$ over $\mathrm{SE}(3)$. We define this time-dependent path $P_t(p\mid O)$ for $t\in[0,1]$, starting from noise $p_0 \sim \mathcal{N}(0,I)$ and evolving to data $p_1 \sim P_{\text{data}}(p\mid O)$, governed by the ODE:
\begin{equation}
\frac{dp}{dt} = u_t(p,O), \quad p(0)\sim\mathcal{N}(0,I).
\end{equation}
Here, $u_t$ represents a time-dependent velocity field parameterized by a neural network $\bm{\Gamma}_{\theta} : \mathbb{R}^{|P|} \times \mathbb{R}^1 \times \mathbb{R}^{3\times N} \rightarrow \mathbb{R}^{|P|}$. To train this network, we utilize the conditional flow-matching objective, which regresses the network output against the ground-truth velocity field~\cite{Lipman2022FlowMF, Ho2022ClassifierFreeDG}:
\begin{equation}
\begin{aligned}
\mathcal{L}_{\text{FM}}(\theta)
&= \mathbb{E}_{t\sim \mathcal{U}(0,1)}
   \mathbb{E}_{p(0)\sim P_{\text{noise}}(p\mid O)}\\
&\quad \mathbb{E}_{p(1)\sim P_{\text{data}}(p\mid O)}
  \left\|
    \bm{\Gamma}_{\theta}(p(t),t\mid O)
    - u_t
  \right\|^2,
\end{aligned}
\end{equation}
where the interpolant is defined as $p(t)=(1-t)p(0) + t\,p(1)$, inducing the straight-line ground-truth velocity $u_t(p(t)\mid p(0),p(1)) = p(1)-p(0)$.

We employ PointNet++~\cite{Qi2017PointNetDH} to extract global geometric features from the point cloud $O$, which are then fused with pose embeddings and time embeddings to predict the velocity. During inference, we reconstruct the pose by numerically integrating the learned ODE via Euler steps:
\begin{equation}\label{eq:flow_integration}
p(1) = p(0) + \int_0^1 \bm{\Gamma}_{\theta}(p(t),t\mid O^*)\, dt.
\end{equation}
By repeating this process $K$ times, we generate a set of $K$ pose candidates. However, while flow matching is adept at covering the distribution, it treats all valid samples equally, lacking a mechanism to identify the optimal pose among the results. This limitation motivates our subsequent reinforcement learning phase.

Standard flow matching integration is deterministic and task-agnostic, it blindly follows the vector field without explicitly minimizing geometric error. To inject the awareness of minimizing errors into the generation process, We propose a paradigm shift by treating the iterative integration steps in Eq.~\eqref{eq:flow_integration} as actions in an MDP. This allows us to refine the generation policy via reinforcement learning, specifically targeting pose accuracy.

\begin{figure}[t]
\centering
\includegraphics[width=0.95\columnwidth]{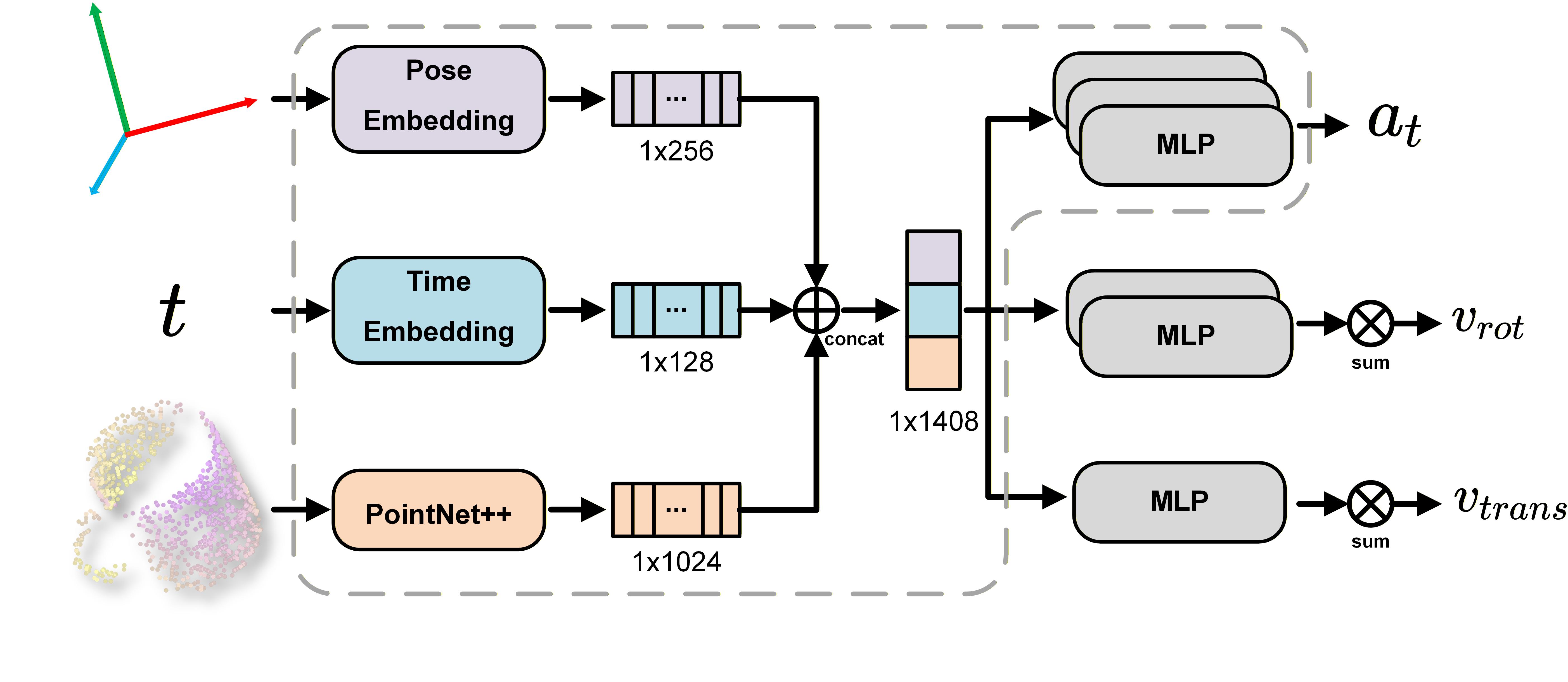}
\caption{
Network architecture of the policy and multi-critic value network. The model processes pose at timestep $t$, timestep embedding, and partial observation point cloud.
}
\label{fig:network_architecture}
\end{figure}

\subsection{PPO Training for Policy Refinement}
Since the policy network $\pi_{\theta'}$ is initialized from the pre-trained flow model, it already resides in a near-optimal region of the parameter space. Extensive exploration is therefore unnecessary, our objective is to fine-tune the generation trajectory for improved geometric accuracy. We model the policy as a gaussian distribution centered $\mu_h=\pi_{\theta'}(s_h)$ on the network output with a small fixed standard deviation $\sigma = 0.2$ for local exploration, sampling actions stochastically as $a_{h+1} \sim \mathcal{N}(\mu_h, \sigma^2 I)$, triggering the state transition:
\begin{equation}
s_{h+1} = \left\{O, \mathbf{u}_{h} \cup \{a_{h+1}\}, (h+1)/H\right\}.
\end{equation}

A critical challenge in 6D pose estimation is the disparity between rotation and translation spaces that they reside on different manifolds and have distinct scales. A single scalar reward would inevitably struggle to balance them. To address this, we introduce a multi-critic value decomposition, as illustrated in Fig.~\ref{fig:network_architecture}, utilizing separate heads for each modality:
\begin{equation}
V_{\phi}(s_h) =
\begin{bmatrix}
V_{\phi}^{\text{rot}}(s_h),
V_{\phi}^{\text{trans}}(s_h)
\end{bmatrix}^T.
\end{equation}

Accordingly, we design specific rewards to penalize deviations from the ground truth, exponentially scaling the errors for sharper signal:
\begin{equation}
\begin{aligned}
r_h^{\text{rot}} &= \exp(-\Delta R_h / \tau_R) + b_h \cdot \mathbb{1}[\Delta R_h < \tau_R], \\
r_h^{\text{trans}} &= \exp(-\Delta T_h / \tau_T) + b_h \cdot \mathbb{1}[\Delta T_h < \tau_T],
\end{aligned}
\end{equation}
where $\Delta R_h$ and $\Delta T_h$ denote the rotation and translation errors between the predicted pose calculated from state $s_h$ and the ground truth pose at step $h$, respectively. The parameters $\tau_R$ and $\tau_T$ control both the sensitivity of the exponential decay and serve as precision thresholds. The indicator function $\mathbb{1}[\cdot]$ activates a constant bonus $b_h = 1$ when the error falls below the corresponding threshold, providing an additional incentive for fine-grained refinement.

With the rotation and translation critics defined, we compute temporal-difference residuals for each head $i \in \{\text{rot}, \text{trans}\}$:
\begin{equation}
\delta_h^i = r_h^i + V_{\phi}^i(s_{h+1}) - V_{\phi}^i(s_h),
\end{equation}
where we adopt the convention $V_{\phi}^i(s_H) = 0$ since no future rewards exist beyond the integration horizon. From these residuals, advantage estimates are computed using Generalized Advantage Estimation (GAE):
\begin{equation}
\hat{A}_h^i = \sum_{l=0}^{H-h} \lambda^l \delta_{h+l}^i,
\end{equation}
where $\lambda$ is the GAE trade-off parameter~\cite{Schulman2015HighDimensionalCC}.

While the critics specialize in disentangling rotational and translational performance, the policy itself remains unified. To this end, we construct a joint advantage signal by equally combining the two estimates,
\begin{equation}
\bar{A}_h = \frac{1}{2}\left( \hat{A}_h^{\text{rot}} + \hat{A}_h^{\text{trans}} \right),
\end{equation}
which is subsequently used to optimize the policy via the clipped PPO objective. And each critic is trained by minimizing a regression loss toward its corresponding advantage target,
\begin{equation}
\mathcal{L}^{V}(\phi) = 
\sum_{i \in \{\text{rot},\text{trans}\}}
\mathbb{E}\!\left[
\big(V_{\phi}^i(s_h)-\, \hat{A}_h^i \big)^2
\right].
\end{equation}

Beyond stabilizing policy optimization, multi-critic value network also learned assessments of pose quality. Rather than being discarded after training, they are retained and later repurposed as confidence estimators in the final pose aggregation stage.

\subsection{Value-Guided Pose Aggregation}

In the final stage, the optimized policy generates $K$ pose candidates through independent trajectories initialized from noise $p_0^{(k)} \sim \mathcal{N}(0,I)$. Each trajectory terminates at state $s_H^{(k)}$, yielding a pose estimate $\hat{p}^{(k)} = \left( \hat{R}^{(k)}, \hat{T}^{(k)} \right).$

Selecting a single optimal pose from these candidates is non-trivial due to geometric ambiguity and partial observability. Instead of relying on likelihood scores or direct averaging, we leverage the learned multi-critic value network as a selector. Specifically, the value network provides disentangled confidence assessments for rotation and translation:
\begin{equation}
V_{\phi}(s^{(k)}) =
\begin{bmatrix}
V_{\phi}^{\text{rot}}(s_{H-1}^{(k)}), \;
V_{\phi}^{\text{trans}}(s_{H-1}^{(k)})
\end{bmatrix}^T,
\end{equation}

We then independently rank the $K$ candidates according to $V_{\phi}^{\mathrm{rot}}$ and $V_{\phi}^{\mathrm{trans}}$, respectively.
For each criterion, the candidates are sorted in descending order, and only the top $\rho$ proportion is retained.
The selected rotations are converted to unit quaternions $\{q_i\}_{i=1}^{\rho K}$. To respect the non-Euclidean structure of $\mathrm{SO}(3)$, we employ a quaternion averaging scheme follow~\cite{Markley2007AveragingQ}:
\begin{equation}
\hat{q}
=
\arg\max_{q \in \mathrm{SO}(3)}
q^\top
\left(
\frac{1}{\rho K}
\sum_{i=1}^{\rho K}
q_i q_i^\top
\right)
q,
\end{equation}
and the resulting quaternion $\hat{q}$ is then converted back to the rotation matrix $\hat{R}$.

Differently, for translation, we aggregate the selected translations using an Euclidean mean: $\hat{T} = \frac{1}{\rho K} \sum_{i=1}^{\rho K} \hat{T}^{(i)}.$ This decoupled, value-guided aggregation strategy effectively handles objects with symmetric geometries where rotational and translational ambiguity may vary independently, resulting in the final robust estimate $\hat{p}=(\hat{R}, \hat{T})$.


\section{Experiment}\label{Experiment}

In this section, we evaluate RFM-Pose on the REAL275, CAMERA~\cite{Wang2019NormalizedOC}, and Omni6DPose~\cite{Zhang2024Omni6DPoseAB} benchmarks. 
We first report overall category-level pose estimation results and efficiency comparisons with representative deterministic and probabilistic methods, followed by per-category analysis on REAL275. 
We further assess the performance of our method on category-level pose tracking and generalization to the Omni6DPose dataset. 
Finally, qualitative results and ablation studies are provided to analyze the effects of RL refinement, sampling strategies, and value-guided aggregation.

\begin{table*}[!htb]
\centering
\caption{Quantitative comparison on REAL275 and CAMERA benchmarks. Best results among probabilistic methods are in \textbf{bold}. $H$ denotes the number of sampling steps. The ``Prior'' denotes whether the method requires category-level shape priors. A dash (``-'') indicates that the corresponding metric was not reported in the original paper.}
\label{tab:nocs_comparison}
\resizebox{\textwidth}{!}{
\begin{tabular}{c|c|cc|cccc|c}
\toprule
\multicolumn{2}{c|}{Method} & Data & Prior &
$5^{\circ}2$cm$\uparrow$ & $5^{\circ}5$cm$\uparrow$ &
$10^{\circ}2$cm$\uparrow$ & $10^{\circ}5$cm$\uparrow$ &
Parameters(M)$\downarrow$ \\
\midrule
& NOCS~\cite{Wang2019NormalizedOC} & RGB-D & $\times$ & - / 32.3 & 9.5 / 40.9 & 13.8 / 48.2 & 26.7 / 64.6 & - \\
& CASS~\cite{Chen2020LearningCS} & RGB-D & $\times$ & 19.5 / - & 23.5 / - & 50.8 / -& 58.0 / - & 47.2 \\
& DualPoseNet~\cite{Lin2021DualPoseNetC6} & RGB-D & $\times$ & 29.3 / 64.7 & 35.9 / 70.7 & 50.0 / 77.2 & 66.8 / 84.7 & 67.9 \\
& SPD~\cite{tian2020shapepriordeformationcategorical} & RGB-D & \checkmark & 19.3 / 54.3 & 21.4 / 59.0 & 43.2 / 73.3 & 54.1 / 81.5 & 18.3 \\
& CR-Net~\cite{Wang2021CategoryLevel6O} & RGB-D & \checkmark & 27.8 / 72.0 & 34.3 / 76.4 & 47.2 / 81.0 & 60.8 / 87.7 & - \\
& SGPA~\cite{Chen2021SGPASP} & RGB-D & \checkmark & 35.9 / 70.7 & 39.6 / 74.5 & 61.3 / 82.7 & 70.7 / 88.4 & - \\

Deterministic
& DPDN~\cite{Lin2022CategoryLevel6O} & RGB-D & \checkmark & 46.0 / - & 50.7 / - & 70.4 / - & 78.4 / - & - \\
\cmidrule{2-9}

& FS-Net~\cite{Chen2021FSNetFS} & D & $\times$ & 19.9 / - & 33.9 / - & - / - & 69.1 / - & 41.2 \\
& GPV-Pose~\cite{Di2022GPVPoseCO} & D & $\times$ & 32.0 / 72.1 & 42.9 / 79.1 & 55.0 / - & 73.3 / 89.0 & - \\
& SAR-Net~\cite{lin2022sarnetshapealignmentrecovery} & D & \checkmark & 31.6 / 66.7 & 42.3 / 70.9 & 50.4 / 75.3 & 68.3 / 80.3 & 6.3 \\
& SSP-Pose~\cite{Zhang2022SSPPoseSS} & D & \checkmark & 34.7 / 64.7 & 44.6 / 75.5 & - / - & 77.8 / 87.4 & - \\
& RBP-Pose~\cite{Zhang2022RBPPoseRB} & D & \checkmark & 38.2 / 73.5 & 48.1 / 79.6 & 63.1 / 82.1 & 79.2 / 89.5 & - \\
\midrule\midrule

& GenPose(H=500)~\cite{zhang2024generative} & D & $\times$ & 52.1 / 79.9 & 60.9 / 84.4 & 72.4 / 84.6 & 84.0 / 89.6 & 4.4 \\

Probabilistic
& Ours(H=20) & D & $\times$ & \textbf{58.4} / \textbf{81.4} & \textbf{65.6} / \textbf{85.1} & \textbf{76.2} / \textbf{86.2} & \textbf{86.3} / \textbf{90.2} & \textbf{4.4} \\
& Ours(H=10) & D & $\times$ & 57.7 / 81.2 & 65.2 / 84.8 & 75.6 / 86.1 & 86.2 / 89.9 & 4.4 \\
& Ours(H=5) & D & $\times$ & 57.1 / 81.1 & 64.9 / 84.5 & 75.0 / 85.7 & 86.1 / 89.7 & 4.4 \\
\bottomrule
\end{tabular}
}
\end{table*}

\begin{figure*}[!htb]
\centering
\includegraphics[width=0.9\textwidth]{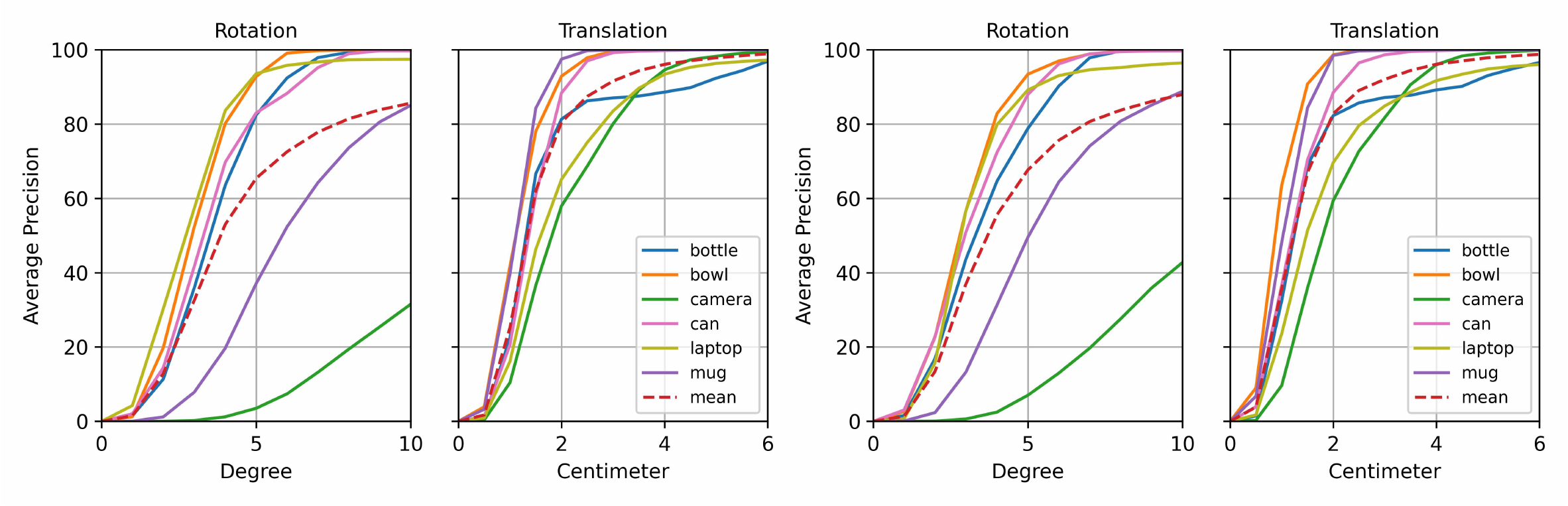}
\caption{
Per-category quantitative comparison with GenPose~\cite{zhang2024generative} on REAL275. The left represents GenPose, while the right represents our approach.
}
\label{fig:AP_compare_quantitative}
\end{figure*}

\begin{table}[!htb]
\centering
\caption{Quantitative results of single-object category-level pose estimation on the REAL275 dataset, where the ``Amb.'' column indicates rotational ambiguity.}
\label{tab:real275_single_compare}
\resizebox{\columnwidth}{!}{
\begin{tabular}{c|c|cc|cc}
\toprule
\multirow{2}{*}{Category} & \multirow{2}{*}{Amb.} & \multicolumn{2}{c|}{$5^{\circ}2$cm} & \multicolumn{2}{c}{$5^{\circ}5$cm}\\ \cmidrule{3-6} 
            &            & GenPose~\cite{zhang2024generative}  & Ours                           & GenPose~\cite{zhang2024generative}  & Ours                           \\ \midrule
camera      &            & 2.9       & \textbf{5.6}                   & 3.2       & \textbf{7.0}                   \\
laptop      & $\times$   & 63.4      & \textbf{66.2}                  & \textbf{91.4}      & 87.2                  \\
average     &            & 32.2      & \textbf{35.9}({\scriptsize{3.7$\uparrow$}})  & \textbf{47.3}      & 47.1({\scriptsize{0.2$\downarrow$}})   \\ \midrule
bottle      &            & 52.6      & \textbf{61.0}                  & 60.9      & \textbf{69.5}                  \\
bowl        &            & 85.4      & \textbf{91.8}                  & 92.6      & \textbf{93.3}                  \\
can         & \checkmark & 72.5      & \textbf{77.6}                  & 80.4      & \textbf{87.0}                  \\
average     &            & 70.2      & \textbf{76.8}({\scriptsize{6.6$\uparrow$}})  & 78.0      & \textbf{83.3}({\scriptsize{5.3$\uparrow$}})  \\ \midrule   
mug         & -          & 35.7      & \textbf{48.5}({\scriptsize{12.8$\uparrow$}})  & 36.4      & \textbf{49.5}({\scriptsize{13.1$\uparrow$}})  \\ \bottomrule
\end{tabular}
}
\end{table}

\begin{table}[!htb]
\centering
\caption{Quantitative results of the category-level pose tracking. Performance comparison on REAL275. Best in \textbf{bold}, second \underline{underlined}.}
\label{tab:pose_tracking_comparison}
\resizebox{\columnwidth}{!}{
\begin{tabular}{@{}lcccccc@{}}
\toprule
Method & Input & Init. & FPS$\uparrow$ & 5°5cm$\uparrow$ & $R_{\text{err}}$(°)$\downarrow$ & $t_{\text{err}}$(cm)$\downarrow$ \\
\midrule
Oracle ICP~\cite{Li2019CategoryLevelAO} & RGBD & GT. & - & 0.7 & 40.3 & 7.7 \\
6-PACK~\cite{Li2019CategoryLevelAO} & RGBD & GT.P. & 14.5 & 33.3 & 16.0 & 3.5 \\
iCaps~\cite{Wang20196PACKC6} & RGBD & Det. & 3.1 & 31.6 & 9.5 & 2.3 \\
CAPTRA~\cite{Li2019CategoryLevelAO} & D & GT.P. & 15.8 & 62.2 & 5.9 & 7.9 \\
CATRE~\cite{Liu2022CATREIP} & D & GT.P. & \textbf{96.6} & 57.2 & 6.8 & \textbf{1.2} \\
GenPose~\cite{zhang2024generative} & D & GT.P. & 21.6 & \underline{71.5} & \textbf{4.2} & \underline{1.5} \\
\midrule
\textbf{Ours} & D & GT.P. & \underline{34.5} & \textbf{75.1} & \underline{5.2} & \textbf{1.2} \\
\bottomrule
\end{tabular}%
}  
\end{table}

\begin{table}[t]
\centering
\caption{Quantitative comparison on Omni6DPose dataset. Best results among probabilistic methods are in \textbf{bold}.}
\label{tab:omni6dpose_comparison}
\resizebox{\columnwidth}{!}{
\begin{tabular}{cc|cc|cccc}
\toprule
\multicolumn{2}{c|}{\multirow{1}{*}{Method}} & \multirow{1}{*}{Data} & \multirow{1}{*}{Prior} & $5^{\circ}2$cm$\uparrow$ & $5^{\circ}5$cm$\uparrow$ & $10^{\circ}2$cm$\uparrow$ & $10^{\circ}5$cm$\uparrow$ \\
\midrule
\multicolumn{1}{c|}{\multirow{4}{*}{Deterministic}} & NOCS~\cite{Wang2019NormalizedOC} & RGB-D & $\times$ & 0.0 & 0.0 & 0.0 & 0.0 \\
\multicolumn{1}{c|}{} & SGPA~\cite{Chen2021SGPASP} & RGB-D & \checkmark & 4.3 & 6.7 & 9.3 & 15.0 \\
\multicolumn{1}{c|}{} & IST-Net~\cite{Lin2022CategoryLevel6O} & RGB-D & $\times$ & 2.0 & 3.4 & 5.3 & 8.8 \\
\multicolumn{1}{c|}{} & HS-Pose~\cite{Zheng2023HSPoseHS} & D & \checkmark & 3.5 & 5.3 & 8.4 & 12.7 \\
\midrule
\multicolumn{1}{c|}{\multirow{3}{*}{Probabilistic}} & GenPose~\cite{zhang2024generative} & D & $\times$ & 6.6 & 9.6 & 13.1 & 19.3 \\
\multicolumn{1}{c|}{} & GenPose++~\cite{Zhang2024Omni6DPoseAB} & RGB-D & $\times$ & 10.0 & 15.1 & \textbf{19.5} & \textbf{29.4} \\
\multicolumn{1}{c|}{} & RFM-Pose(Ours) & D & $\times$ & \textbf{11.8} & \textbf{20.9} & 13.9 & 25.4 \\
\bottomrule
\end{tabular}
}
\end{table}

\begin{figure*}[!htb]
\centering
\includegraphics[width=0.8\textwidth]{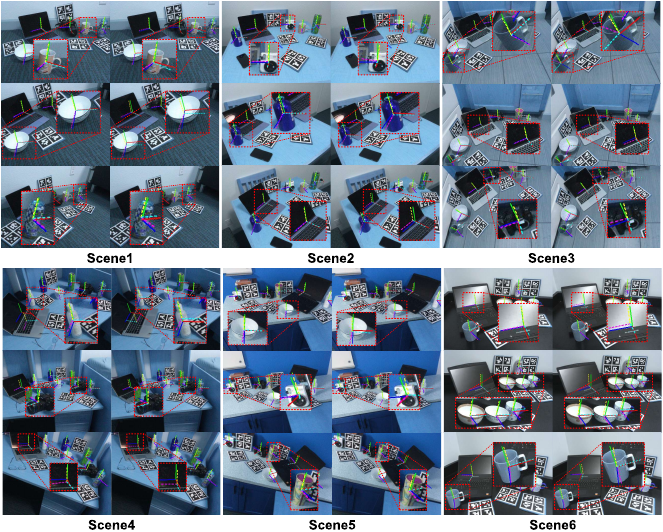}
\caption{
Qualitative comparison between our method and GenPose on six representative scenes from the REAL275 test set. Left is our method, right is the baseline method for reference}
\label{fig:scene1_to_6}
\end{figure*}

\begin{figure*}[!htb]
\centering
\includegraphics[width=0.85\textwidth]{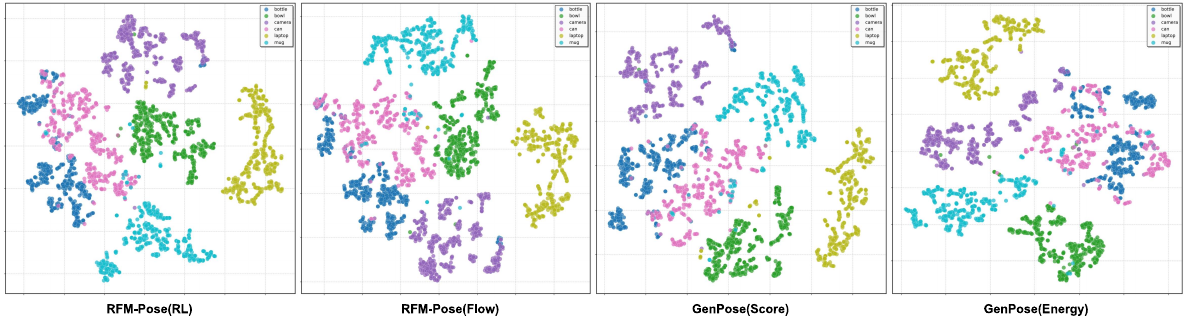}
\caption{
t-SNE visualization of point cloud feature distributions across different methods. Each point represents a feature vector colored by object category.}
\label{fig:tsne_analysis}
\end{figure*}

\begin{figure}[!htb]
\centering
\includegraphics[width=0.9\columnwidth]{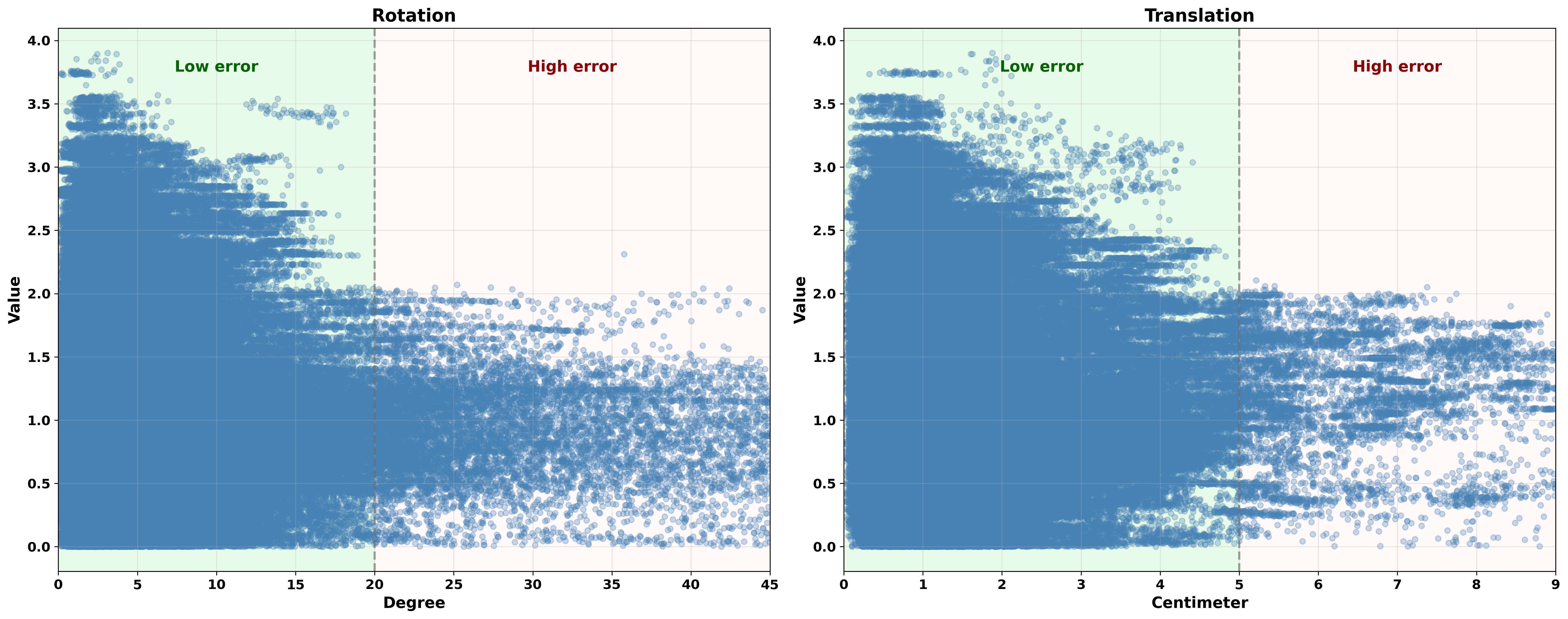}
\caption{
Scatter plots of predicted value scores versus actual pose errors for rotation (left) and translation (right).}
\label{fig:value_error_scatter}
\end{figure}

\begin{figure*}[!htb]
\centering
\includegraphics[width=0.9\textwidth]{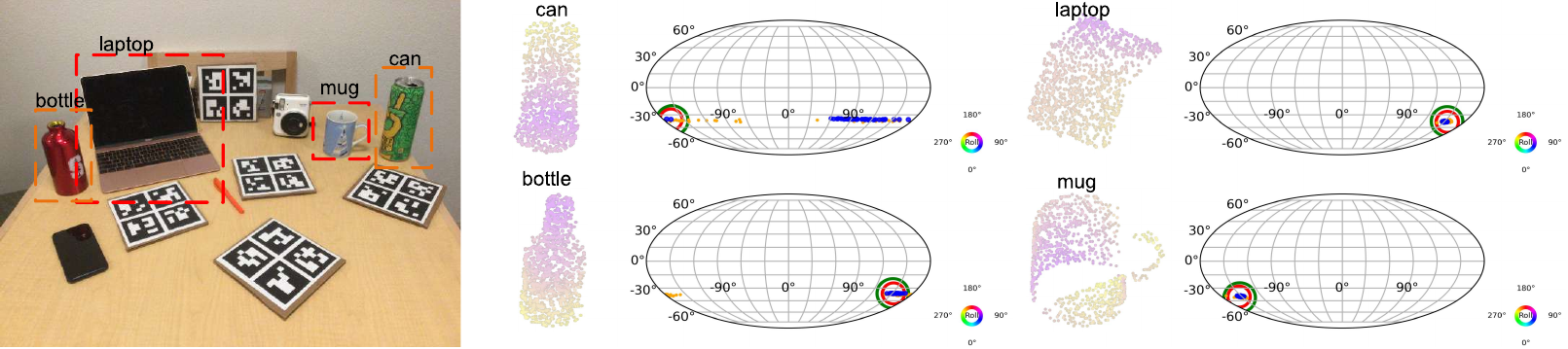}
\caption{
SO(3) rotation distributions for objects with relatively symmetric and clearly asymmetric geometries across three scenes.}
\label{fig:so3_analysis}
\end{figure*}

\begin{table}[!htb]
\centering
\caption{Ablation study on sampling count $K$ and selection ratio $\rho$}
\label{tab:ablation_k_delta}
\begin{tabular}{c|ccccc}
\toprule
\diagbox{$K$}{$\rho$} & 20\% & 40\% & 60\% & 80\% & 100\% \\ 
\midrule
10 & 75.2 & 75.7 & 75.9 & 75.8 & 75.1 \\
30 & 75.5 & 75.9 & 75.9 & 76.0 & 75.3 \\
50 & \textbf{75.7} & \textbf{76.0} & \textbf{76.2} & \textbf{76.1} & \textbf{76.0} \\ 
\bottomrule
\end{tabular}
\end{table}

\begin{table}[!htb]
\centering
\caption{Ablation study on ranking strategies.}
\label{tab:ablation_ranking}
\begin{tabular}{cc|cccc}
\toprule
Ranking & Mean & $5^{\circ}2$cm & $5^{\circ}5$cm & $10^{\circ}2$cm & $10^{\circ}5$cm \\ 
\midrule
Random & $\times$ & 50.8 & 59.3 & 71.3 & 83.1 \\
Random & \checkmark & 57.8 & 64.6 & 76.0 & 85.8 \\
Value & \checkmark & \textbf{58.4} & \textbf{65.6} & \textbf{76.2} & \textbf{86.3} \\
GT (oracle) & \checkmark & 64.2 & 70.8 & 79.2 & 88.5 \\ 
\bottomrule
\end{tabular}
\end{table}

\begin{table}[!htb]
\centering
\caption{Ablation study on flow matching, RL refinement, and aggregation strategies. Results are mAP (\%) on REAL275.}
\label{tab:ablation_flow_rl}
\begin{tabular}{c|cc|cc}
\toprule
\multirow{2}{*}{Method} & \multicolumn{2}{c|}{Mean Aggregation} & \multicolumn{2}{c}{Value Aggregation} \\
\cmidrule(lr){2-3} \cmidrule(lr){4-5}
& $5^{\circ}2$cm & $10^{\circ}5$cm & $5^{\circ}2$cm & $10^{\circ}5$cm \\
\midrule
Flow & 45.9 & 81.2 & 49.8 & 81.9 \\
RL & 57.8 & 85.8 & \textbf{58.4} & \textbf{86.3} \\
\bottomrule
\end{tabular}
\end{table}

\subsection{Experimental Setup}

\textbf{Datasets.}
We evaluate on three benchmarks: CAMERA and REAL275, and Omni6DPose. CAMERA is a synthetic dataset with 275K training and 25K testing images across six categories, namely bottle, bowl, camera, can, laptop, and mug. REAL275 contains real RGB-D data with 4.3K training images from 7 scenes and 2.75K testing images from 6 scenes for the same categories. Omni6DPose provides large-scale RGB-D data with broader object diversity. We follow the official splits and protocols for all datasets.

\textbf{Metrics.}
We report mean Average Precision (mAP) at four threshold combinations: $5^\circ$\,2cm, $5^\circ$\,5cm, $10^\circ$\,2cm, and $10^\circ$\,5cm. Following the standard REAL275/CAMERA evaluation protocol~\cite{Wang2019NormalizedOC}, categories with strong rotational ambiguity such as bottle, bowl, and can are evaluated with symmetry-aware rotation error. For mug, the effective symmetry depends on handle visibility, and the symmetry-aware criterion is applied when the handle is largely occluded.

\textbf{Implementation.}
We use Mask R-CNN~\cite{2017Mask} for instance segmentation and adopt the backbone architecture from GenPose~\cite{zhang2024generative}. Input point clouds are sampled to 1024 points for NOCS datasets. Training consists of two stages, where the flow-matching model training on each dataset, followed by reinforcement learning fine-tuning initialized from stage one. We apply data augmentation including random rotation, scaling, and point jittering following FS-Net~\cite{Chen2021FSNetFS}. All results use 20 sampling steps and 50 trials per instance unless otherwise noted. Experiments are conducted on an NVIDIA RTX 4090 GPU with batch size 256 using PyTorch.

\subsection{Quantitative Evalution}

\subsubsection{\textbf{Comparison with representative methods}}
Table~\ref{tab:nocs_comparison} summarizes the quantitative comparison on REAL275 and CAMERA benchmarks. We compare against both deterministic and probabilistic methods. Higher values indicate better performance for pose accuracy metrics, while lower values are preferred for model parameters. Our method achieves 58.4\% mAP on REAL275 under the strict $5^{\circ}2$cm metric with only 20 sampling steps, outperforming previous work~\cite{zhang2024generative} by 6.3\% while using 25$\times$ fewer integration steps. Under $5^{\circ}5$cm, $10^{\circ}2$cm, and $10^{\circ}5$cm, we observe consistent improvements of 4.7\%, 3.8\%, and 2.3\%, respectively. On the CAMERA dataset, our approach also achieves favorable results across all metrics. Notably, even with significantly reduced sampling steps ($H$=10 or $H$=5), RFM-Pose maintains competitive accuracy, demonstrating the excellent efficiency-accuracy trade-off enabled by our RL-refined flow matching framework.

\subsubsection{\textbf{Per-category evaluation on REAL275}}
Fig.~\ref{fig:AP_compare_quantitative} and Table~\ref{tab:real275_single_compare} report per-category results on REAL275. We observe consistent gains for both clearly asymmetric objects and objects with stronger rotational ambiguity, with average improvements of 6.6\% and 5.3\% under $5^{\circ}2$cm and $5^{\circ}5$cm, respectively. Notably, mug exhibits the largest improvement.

\subsubsection{\textbf{Category-level pose tracking on REAL275}}
Category-level pose tracking~\cite{9792223} aims to continuously estimate object poses across video frames, which is essential for robotic manipulation and augmented reality applications. Table~\ref{tab:pose_tracking_comparison} presents pose tracking results on REAL275. Our method achieves 75.1\% accuracy under $5^{\circ}5$cm, outperforming GenPose by 3.6\% while operating at 34.5 FPS (60\% faster than GenPose's 21.6 FPS). Rotation and translation errors are 5.2° and 1.2 cm, respectively, demonstrating favorable accuracy-speed balance.

\subsubsection{\textbf{Generalization to Omni6DPose}}
On the challenging Omni6DPose dataset (Table~\ref{tab:omni6dpose_comparison}), our method achieves 11.8\% and 20.9\% mAP under $5^{\circ}2$cm and $5^{\circ}5$cm, outperforming GenPose by 5.2\% and 11.3\%, respectively. Under strict thresholds, our approach demonstrates effective generalization to diverse object categories beyond the standard NOCS benchmark. We note that GenPose++ employs additional RGB features, which may contribute to its advantage under relaxed metrics.

\subsection{Qualitative Evaluation}

\subsubsection{\textbf{Qualitative comparison on REAL275 scenes}}
Fig.~\ref{fig:scene1_to_6} presents a qualitative evaluation on six scenes from the REAL275 test set, comparing the proposed method against GenPose. Each scene comprises multiple object instances captured from different viewpoints. Ground-truth poses are represented by solid axes (red, green, blue), while predicted poses are indicated by dashed axes (yellow, purple, light-blue). The proposed method (left column) consistently achieves closer alignment with ground-truth orientations compared to GenPose (right column), as visualized by the reduction in angular deviation between the predicted coordinate system and the ground truth coordinate system. Across all scenes, our approach maintains stable predictions under varying conditions.

\subsubsection{\textbf{Representation analysis with t-SNE}}
Fig.~\ref{fig:tsne_analysis} visualizes point cloud feature distributions using t-SNE~\cite{Maaten2008VisualizingDU}. From left to right, we show RFM-Pose after PPO refinement, RFM-Pose before RL fine-tuning, GenPose baseline, and GenPose with energy-based refinement. RFM-Pose after PPO refinement the most compact and well-separated clusters. Comparing RFM-Pose(Flow) and RFM-Pose(RL) demonstrates the effectiveness of RL refinement, while flow matching learns reasonable clustering, RL further tightens clusters and reduces inter-category overlap. The reward signal measuring rotation/translation errors provides valuable supervision for joint optimization of feature extraction and pose generation. Compared with GenPose variants, our approach shows superior feature organization with less scattered distributions and clearer category boundaries.

\subsection{Ablation Study}

\subsubsection{\textbf{Sampling count and selection ratio}}
Table~\ref{tab:ablation_k_delta} investigates effects of sampling count $K$ and selection ratio $\rho$ on REAL275 under $10^{\circ}5$cm. Increasing $K$ consistently improves performance, with $K=50$ achieving best results. The optimal $\rho=60\%$ balances retaining high-quality hypotheses while filtering outliers. We adopt $K=50$ and $\rho=60\%$ as default.

\subsubsection{\textbf{Ranking strategies for hypothesis selection}}
Table~\ref{tab:ablation_ranking} compares the ranking strategies for sampling and summarizing the top 60\% candidates on the REAL275 dataset. Selection based on the learned value network achieves 58.4\% at $5^{\circ}2$cm, outperforming both random selection and single random sample. The gap between value network and oracle GT ranking suggests potential for improved value estimation, though this gap narrows to 2.2\% under relaxed metrics.

\subsubsection{\textbf{Value-error correlation analysis}}
Fig.~\ref{fig:value_error_scatter} shows the correlation between predicted values and actual errors, where each point represents a pose candidate and color indicates error magnitude. High-value regions consistently correspond to low errors, while low-value regions exhibit greater variance. This asymmetric distribution validates our value-guided aggregation strategy, as high-value scores reliably identify accurate poses, enabling effective hypothesis selection without ground-truth supervision at inference time.

\subsubsection{\textbf{Effect of RL refinement and aggregation strategies}}
Table~\ref{tab:ablation_flow_rl} isolates contributions of RL refinement and aggregation strategies, where ``Flow'' denotes the flow-matching baseline before fine-tuning, ``RL'' indicates the refined policy, ``Mean'' denotes uniform averaging, and ``Value'' denotes selection guided by the learned value network. The flow-matching baseline achieves 45.9\% under $5^{\circ}2$cm with mean aggregation, while RL refinement substantially improves this to 57.8\%, demonstrating that explicit error rewards successfully steer sampling toward higher-quality regions. Additionally, value network-guided aggregation provides consistent gains over mean aggregation for both Flow and RL settings. The combination of RL refinement and value network-guided aggregation achieves the best performance, validating the complementary benefits of our two components.

\subsubsection{\textbf{SO(3) distribution analysis}}
Fig.~\ref{fig:so3_analysis} visualizes SO(3) rotation distributions for objects with varying degrees of geometric symmetry. Blue points denote RL-refined rotation candidates, yellow points indicate flow-matching baseline rotation candidates, green circles mark ground-truth rotations, and red circles show the final selected predictions. Compared with the flow-matching baseline, the RL-refined policy produces noticeably more concentrated candidates around the ground-truth neighborhood, while the value-guided selection consistently identifies high-quality candidates. The improvement is particularly pronounced for objects with stronger rotational ambiguity induced by relatively symmetric shapes, where the baseline exhibits dispersed distributions but RL refinement effectively tightens the candidate clusters.

\FloatBarrier

\section{Conclusion}\label{Conclusion}


We proposed RFM-Pose, which integrates flow matching with reinforcement learning by reformulating sampling as an MDP for policy-guided refinement and value-based hypothesis ranking. Experiments demonstrate favorable accuracy-efficiency trade-offs over generative baselines. Future work will explore RGB cues and broader object categories.

\end{document}